\documentclass[conference]{IEEEtran}
\IEEEoverridecommandlockouts

\usepackage{cite}
\usepackage{amsmath,amssymb,amsfonts}
\usepackage{algorithmic}
\usepackage{graphicx}
\usepackage{textcomp}
\usepackage{xcolor}
\usepackage{multirow}
\usepackage{booktabs}
\usepackage{enumitem}
\usepackage{float}  
\usepackage{pifont}
\usepackage{listings}
\usepackage{subcaption}
\usepackage{tabularx}
\usepackage[hyphens]{url} 
\usepackage{algorithm}

\newcommand{\cmark}{\ding{51}} 
\newcommand{\xmark}{\ding{55}} 

\def\BibTeX{{\rm B\kern-.05em{\sc i\kern-.025em b}\kern-.08em
    T\kern-.1667em\lower.7ex\hbox{E}\kern-.125emX}}
\begin{document}

\title{AceParse: A Comprehensive Dataset with Diverse Structured Texts for Academic Literature Parsing
}


\author{{\begin{tabular}{c}{Huawei Ji}$^1$, Cheng Deng$^1$, Bo Xue$^1$, Zhouyang Jin$^1$, Jiaxin Ding$^{1*}$\thanks{*Corresponding author. This work was supported by the National Natural Science Foundation of China (grant no.61960206002,62020106005, 62432002, 62202299).}, Xiaoying Gan$^1$,\\ {Luoyi Fu$^1$, Xinbing Wang$^1$, Chenghu Zhou$^2$}\end{tabular}} \\ {$^1$Shanghai Jiao Tong University, Shanghai, China} \\
{$^2$IGSNRR, Chinese Academy of Sciences, Beijing, China}
}

\maketitle

\begin{abstract}
With the development of data-centric AI, the focus has shifted from model-driven approaches to improving data quality. Academic literature, as one of the crucial types, is predominantly stored in PDF formats and needs to be parsed into texts before further processing.
However, parsing diverse structured texts in academic literature remains challenging due to the lack of datasets that cover various text structures.
In this paper, we introduce AceParse, the first comprehensive dataset designed to support the parsing of a wide range of structured texts, including formulas, tables, lists, algorithms, and sentences with embedded mathematical expressions. 
Based on AceParse, we fine-tuned a multimodal model, named AceParser, which accurately parses various structured texts within academic literature.
This model outperforms the previous state-of-the-art by 4.1\% in terms of F1 score and by 5\% in Jaccard Similarity, demonstrating the potential of multimodal models in academic literature parsing.
Our dataset is available at \url{https://github.com/JHW5981/AceParse}.

\end{abstract}


\section{Introduction}
\label{sec:intro}

The increasing focus on data-centric AI has shifted the emphasis from model to the critical role of data in advancing AI technologies~\cite{zha2023data, deng2024k2}. In this paradigm, the quality, diversity, and representativeness of data play a pivotal role in determining the success of AI systems. High-quality data are essential for training advanced models, particularly in domains requiring a deep understanding of complex information. Academic literature is a valuable data source due to its rich scientific content~\cite{deng2021gakg}, but it is often stored in non-machine-readable formats like PDFs, necessitating effective parsing techniques to extract their information.

Academic literature typically features a blend of structured content, including tables, formulas, lists, and algorithms, all of which work together to convey scientific insight. Parsing literature presents several key challenges. First, OCR-based methods~\cite{smith2007overview, du2020pp, hegghammer2022ocr} often focus on character recognition, leading to loss of structural information. These methods lack the ability to preserve the hierarchical and relational structure necessary for accurately understanding scientific content. Second, while modular approaches~\cite{opendatalab_mineru, lo2023papermage} can handle predefined content types like tables and formulas, they struggle with complex structures like algorithms and lists. Moreover, the lack of integration between modules often results in inconsistent outputs for interconnected content types. Existing end-to-end parsing models, such as Nougat~\cite{blecher2023nougat}, are often trained on narrow proprietary datasets with a limited diversity of structured content, restricting their ability to generalize effectively across diverse structures. Finally, existing open-source datasets remain limited to character-level parsing~\cite{pfitzmann2022doclaynet, li2020docbank} or focus on specific content types like tables or formulas~\cite{li2020tablebank, deng2017image}, which does not cover the full diversity of structured elements present in academic documents. This gap highlights the pressing need for datasets that encompass a more comprehensive range of structured content to support advanced parsing techniques.

\begin{figure*}[!t]
\centering
\includegraphics[width=\textwidth]{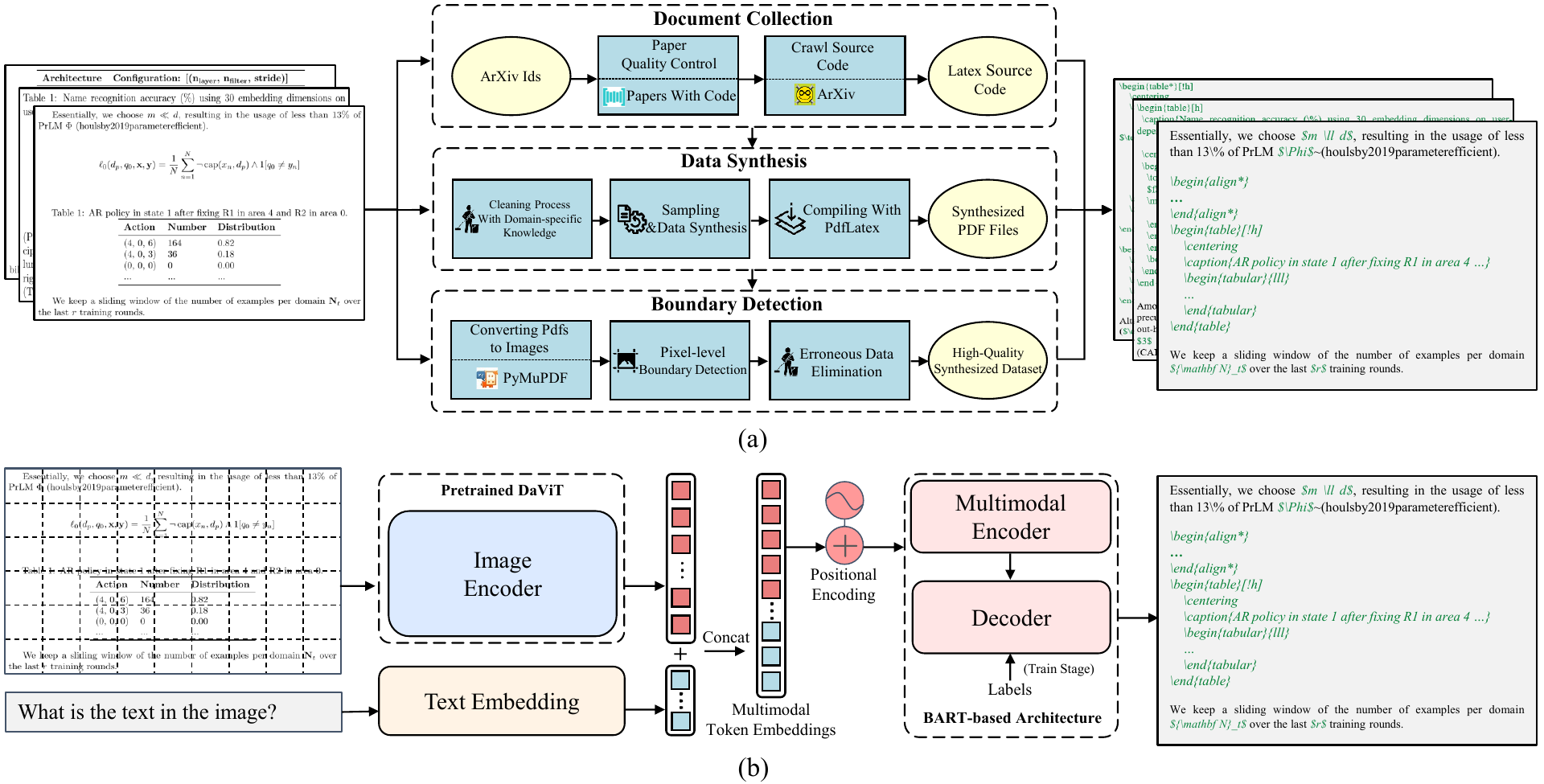}
\captionsetup{skip=0pt}  
\caption{(a) \textbf{The construction process of the AceParse dataset} consists of three stages: document collection, data synthesis, and boundary detection. (b) \textbf{The network architecture of AceParser.} Visual token embeddings and text token embeddings are concatenated into multimodal token embeddings, which are then processed by a BART-based multimodal encoder-decoder.}
\label{dataset_and_model}
\end{figure*}

To advance the unified parsing of academic literature containing various types of structured texts, we propose AceParse. Unlike previous datasets~\cite{pfitzmann2022doclaynet, li2020docbank} that either provide only character-level parsing results or focus on a single type of structured text~\cite{li2020tablebank, deng2017image}, AceParse encompasses a broad range of structured texts, including formulas, tables, lists, algorithms and sentences with
embedded mathematical expressions. This comprehensive coverage addresses the limitations of existing datasets, enabling more versatile parsing solutions. Moreover, these texts are annotated using the LaTeX markup language to accurately describe their structure. As far as we know, AceParse is the first open-source dataset specifically designed for handling diverse structured content in academic literature parsing. A comparison of different datasets is provided in Table~\ref{dataset_comparison}. Based on AceParse, we propose AceParser, a multimodal academic literature parsing model that adopts the network architecture referenced from Florence2~\cite{xiao2024florence}. As shown in Table~\ref{method_comparison}, our model is compared with other existing methods.

The main contributions of this paper are as follows:

\begin{itemize}[itemsep=2pt, parsep=0pt, left=0.8em]
    \item We introduced AceParse, the first comprehensive, open-source dataset designed for parsing diverse structured texts in academic literature.
    \item We fine-tuned a multimodal model to develop AcePa-
    rser, an end-to-end structured text parsing method capable of generating structured text in markup languages.
    \item We systematically compare the performance of current parsing methods and provide an extensive overview of existing parsing datasets, aiming to serve as a reference for the document parsing community.
\end{itemize}

\section{METHODOLOGY}
\label{sec:methodlogy}

\subsection{Dataset Construction}
\label{ssec:dataset_construction}
Existing academic parsing datasets~\cite{blecher2023nougat} rely on a PDF page matching data construction mechanism. Still, the inherent limitations in the page matching model often lead to the loss of structured text at the end of pages. Moreover, this approach requires the collection of a large volume of academic literature, creating substantial challenges for scalability. To address these issues, we use a data synthesis approach. By randomly combining structured texts extracted from source code to generate new LaTeX code and then compiling it, we obtain high-quality image-annotation pairs. Additionally, the randomness in sampling allows us to develop a large volume of data from a small amount of source code. The dataset is built upon three key dimensions, as illustrated in Fig.~\ref{dataset_and_model}(a):

\begin{figure*}[!t]
\centering
\includegraphics[width=\textwidth]{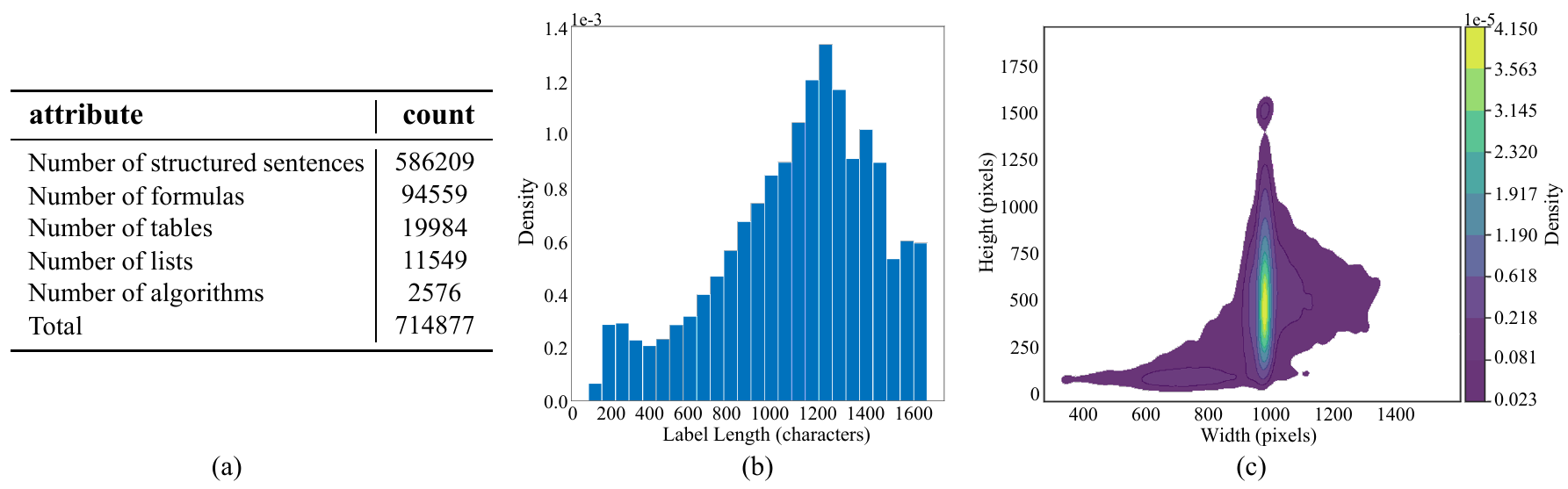}
\captionsetup{skip=0pt}  
\caption{\textbf{Statistics of the AceParse dataset.} (a) The number of structured items comprising AceParse, including the number of structured sentences with embedded formulas, etc. (b) Frequency histogram indexed by the label character length. (c) Joint kernel density estimation of image dimensions.}
\label{dataset_statistics}
\end{figure*}

\begin{table}[!t]
    \centering
    \caption{Comparison of different academic literature parsing datasets. "Modality" indicates the type of content the dataset is designed for. "Structured" refers to whether the labels are represented using markup languages. *AceParse includes various structured texts such as tables, formulas, lists, algorithms, etc.}
    \small
    \begin{tabular}{lccc}
    \toprule
         \textbf{Dataset}& \textbf{Size} & \textbf{Modality}& \textbf{Structured} \\ \midrule
        DocLayNet~\cite{pfitzmann2022doclaynet} & 80k & Text & \xmark \\
        DocBank~\cite{li2020docbank} & 500k & Text & \xmark \\
        TableBank~\cite{li2020tablebank} & 417k & Table & \cmark \\
        IM2LATEX~\cite{deng2017image} & 100k & Formula & \cmark \\ 
        AceParse (Ours) & 500k & Text+Table+Formula* & \cmark \\
        \bottomrule
    \end{tabular}
\label{dataset_comparison}
\end{table}

\textbf{Document Collection:} Using ArXiv IDs listed in Papers with Code following~\cite{hu2023mplug}, we collected 10,000 open-access LaTeX source files from 102 subfields within computer science on ArXiv. During this process, we developed custom parsing scripts to address the structural and formatting inconsistencies of LaTeX files across various subfields. These scripts were specifically designed to normalize differing LaTeX conventions and ensure consistent content extraction.

\textbf{Data Synthesis:} We applied a combination of rule-based techniques, leveraging our domain-specific knowledge of academic writing conventions and LaTeX syntax to clean the source code. To avoid issues such as inconsistent citation formats, overly complex or redundant user-defined commands, and non-standard sectioning in the LaTeX files, we developed custom parsing rules to filter out irrelevant content, normalize references and citations, and standardize or replace non-standard commands. Following this cleaning step, we extracted diverse structured texts, specifically focusing on sentences with embedded structures, formulas, tables, lists, and algorithms while excluding plain text sentences. This approach allowed us to collect over 700,000 structured items, as shown in Figure~\ref{dataset_statistics}(a).

When randomly sampling and combining these items to synthesize new LaTeX files, one of the main challenges was ensuring that the files would compile successfully despite the randomness of the content. Since the combination of different structures, such as formulas, tables, and custom commands, can easily result in compilation errors, we needed to implement strict syntax checks to catch potential issues before the final compilation stage. Additionally, user-defined commands from different sources often conflicted or overlapped, leading to formatting inconsistencies. To resolve this, we introduced a pre-processing step to sanitize and harmonize conflicting commands. After these adjustments, the synthesized TeX files were successfully compiled into PDFs using \texttt{pdflatex}\footnote{\url{https://www.tug.org/applications/pdftex/}}, ensuring that all structures were rendered correctly despite the randomized content.

\begin{table}[!t]
    \centering
    \caption{Comparison of different academic literature parsing methods. "Params" refers to the number of model parameters. "Paradigm" describes whether the model uses a modular approach. "Structured" indicates whether the method can parse structured texts.}
    \begin{tabular}{lccc}
    \toprule
         \textbf{Method}& \textbf{Params} & \textbf{Paradigm}& \textbf{Structured} \\ \midrule
        Tesseract~\cite{smith2007overview} & 5.1M & End-to-End & \xmark \\
        PPOCR~\cite{du2020pp} & 10M & Modular & \xmark \\
        Pix2Text & 84M & Modular & \cmark \\ 
        Mineru~\cite{opendatalab_mineru} & 1.5G & Modular & \cmark \\
        Nougat~\cite{blecher2023nougat} & 350M & End-to-End & \cmark \\
        AceParser (Ours) & 270M & End-to-End & \cmark \\
        \bottomrule
    \end{tabular}
\label{method_comparison}
\end{table}

\textbf{Boundary Detection:} We used a pixel-level boundary detection method to accurately extract literature images. PDFs were first converted to images using \texttt{PyMuPDF}\footnote{\url{https://github.com/pymupdf/PyMuPDF}}. By identifying the corners of text areas, we cropped the images to focus on relevant content. Heuristic rules were then applied to detect page boundaries and discard samples with irregular layouts or distorted content, ensuring cleaner and more precise image extractions for downstream tasks.

\begin{figure*}[!t]
\centering
\includegraphics[width=\textwidth]{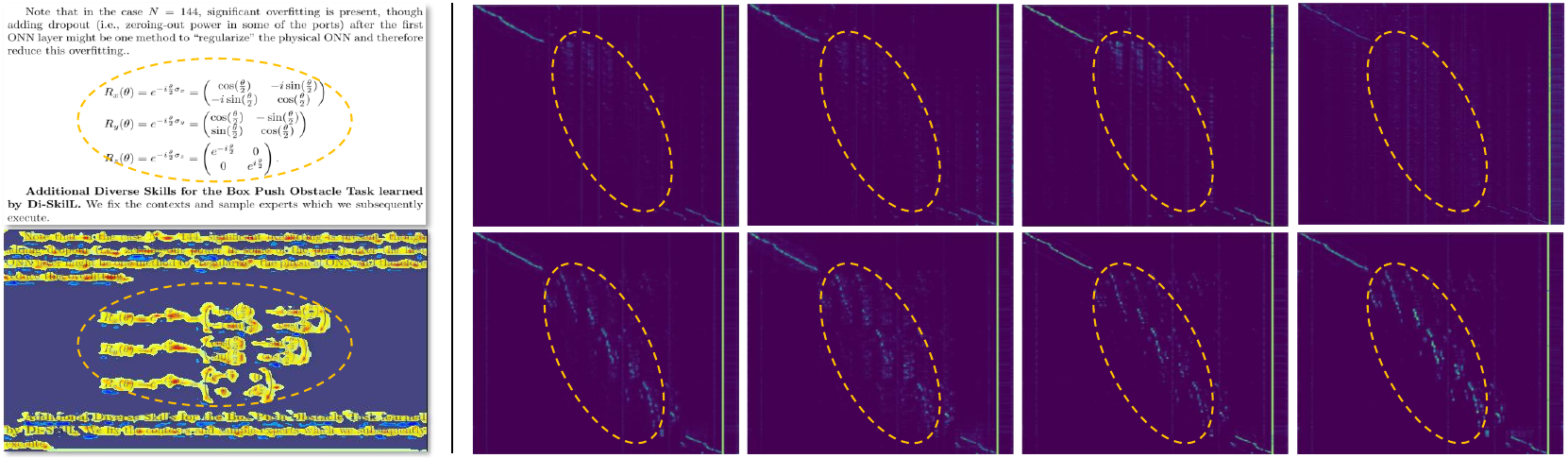}
\captionsetup{skip=5pt}  
\caption{\textbf{Feature map and cross-attention matrices} within the AceParser model. The matrices displayed on the right represent cross-attention from four layers of AceParser, with the top row showing attention before training and the bottom row after training. Structured text locations are highlighted with yellow ellipses.}
\label{mechanism}
\end{figure*}  

Fig. \ref{dataset_statistics} details the statistical attributes of the AceParse dataset. Each synthesized document image contains annotation text centered at 1,107 characters, matching the model's processing capacity. As shown in Figure \ref{dataset_statistics}(c), the majority of images have dimensions centered at 974 $\times$ 493 pixels, with median values for width and height being 974 and 493 pixels, respectively. These dimensions ensure a balance between resolution and file size, optimizing both training efficiency and content clarity.

\subsection{AceParser Network Architecture}
\label{ssec:model}

The proposed AceParser Network is fine-tuned based on the architecture of Florence-2~\cite{xiao2024florence}, as illustrated in Fig.\ref{dataset_and_model}(b). Florence-2 is a robust multi-task multimodal model pre-trained on 5 billion data instances and equipped with OCR capabilities, but it lacks the ability to parse structured texts. The model comprises a vision encoder, DaViT\cite{ding2022davit}, and a multimodal encoder-decoder based on BART~\cite{lewis2019bart}. A document image, denoted as $\mathbf{I} \in \mathbb{R}^{H \times W \times 3}$, is divided into patches and embedded into visual token embeddings $\mathbf{V} \in \mathbb{R}^{N_v \times d}$ by the vision encoder, where $N_v$ represents the number of visual tokens, and $d$ denotes the dimension of the hidden layers. The task prompt is also embedded into text token embeddings $\mathbf{T} \in \mathbb{R}^{N_t \times d}$ by the text embedding layer, where $N_t$ represents the number of text tokens. By concatenating these visual and text token embeddings and applying positional encoding, we obtain the multimodal token embeddings $\mathbf{X} \in \mathbb{R}^{(N_v+N_t) \times d}$, which serve as input to the multimodal encoder. Using teacher forcing and an autoregressive loss during training, we can fine-tune AceParser by providing the annotations as input to the decoder:

\[
\mathcal{L} = - \sum_{t=1}^{T} \log P(y_t | y_{1:t-1}, x)
\]
where \( y_t \) is the actual token at time step \( t \), \( y_{1:t-1} \) are the previous tokens generated by the decoder, and \( x \) represents the input sequence (e.g., the annotations).

\begin{table}[!t]
    \centering
    \small
    \caption{Comparison of different parsing methods using various evaluation metrics. Metric abbreviations are used for clarity, such as LD for Levenshtein Distance.}
    \begin{tabular}{l|ccccc}
    \toprule
         \textbf{Method}& \textbf{LD} $\mathbf{\downarrow}$ & \textbf{BLEU} $\mathbf{\uparrow}$& \textbf{F1} $\mathbf{\uparrow}$& \textbf{JS}$ \mathbf{\uparrow}$& \textbf{Time} \\ \midrule
        Tesseract~\cite{smith2007overview} & 0.52 & 19.3 & 51.3 & 37.2 & \textbf{1.79}\\
        PPOCR~\cite{du2020pp} & 0.53 & 18.4 & 53.4 & 39.4 & 6.26\\
        Pix2Text & 0.43 & 33.6 & 62.6 & 47.2 & 2.47\\
        Mineru~\cite{opendatalab_mineru} & 0.39 & 45.6 & 68.2 & 53.4 & 984.9 \\
        Nougat~\cite{blecher2023nougat} & 0.43 & 44.9 & 68.0 & 53.4 & 11.24 \\
        AceParser (Ours) & \textbf{0.34} & \textbf{50.2} & \textbf{72.3} & \textbf{58.4} & 5.92\\
        Improvements & \textcolor{red}{+0.05} & \textcolor{red}{+4.6} & \textcolor{red}{+4.1} & \textcolor{red}{+5.0} & \textcolor{blue}{-4.13} \\
        \bottomrule
    \end{tabular}
\label{result_comparison}
\end{table}

\section{EXPERIMENTS AND DISCUSSIONS}
\label{sec:results}

\subsection{Experimental Setup}
\label{ssec:setup}

The AceParse dataset is divided into training, validation, and test sets with a ratio of 8:1:1. All comparison results are reported based on the test set. The AceParser model is initialized with pre-trained weights from Florence-2~\cite{xiao2024florence} and trained with the AdamW optimizer, using a learning rate of $1 \times 10^{-5}$ and a linear learning rate schedule, which includes a 10\% warm-up phase. Training is performed on four NVIDIA GeForce RTX 3090 GPUs with a batch size of 8.

\subsection{Comparison with Different Parsing Methods}
\label{ssec:comparison}

As shown in Table 3, we compared various academic literature parsing methods on the AceParse dataset using evaluation metrics including Levenshtein Distance~\cite{levenshtein1966binary}, BLEU~\cite{papineni2002bleu}, F1-score, and Jaccard Similarity~\cite{schutze2008introduction}. Methods that are not structure-aware, such as Tesseract~\cite{smith2007overview} and PPOCR~\cite{du2020pp}, generally showed lower performance, due to their limited ability to handle markup languages. Among structure-aware methods, modular approaches like Pix2Text and Mineru~\cite{opendatalab_mineru} were also less competitive, attributed to error accumulation across modules. Our end-to-end method achieved the best parsing results, with a 4.1-point improvement in the F1 score and a 5-point improvement in Jaccard similarity. However, a current limitation of our approach is its relatively slow parsing speed of 5.92 seconds per sample, which will be a focus for future optimization.

\subsection{Case Study}
\label{ssec:case_study}

We provide a case study where AceParser parses an academic document image containing complex formulas, as shown in Figure~\ref{mechanism}. The two images on the left show the original image and the feature map extracted from the image encoder. It can be observed that the image encoder captures not only the areas with plain text but also focuses on the special symbols and structures within the formulas. The images on the right illustrate the cross-modality attention matrices before and after training. These matrices capture the relationships between the input tokens (both visual and textual) and the corresponding parsed output tokens, highlighting how the model aligns visual and textual information to achieve accurate parsing results. We observe a substantial rise in attention scores within the formula region, suggesting that the model, which initially had difficulty parsing structured text, significantly improved its ability to do so after being trained on AceParse.

\section{CONCLUSIONS}
\label{sec:conclusions}

In this paper, we introduced AceParse, the first comprehensive, open-source academic literature parsing dataset containing multiple types of structured texts, addressing the issue of the lack of diverse structured content in previous datasets. The dataset includes 500k parsed document pairs, annotated in LaTeX, designed to teach models how to represent diverse structured texts using markup language. Based on this dataset, we trained an end-to-end model, AceParser, which achieved state-of-the-art parsing performance. Our work establishes a foundation for academic literature parsing and the development of end-to-end parsing models. In future work, we plan to enhance dataset quality, increase document length, and explore smaller models to improve parsing speed.

\vfill\pagebreak
\bibliographystyle{IEEEtran}
\bibliography{refs}

\end{document}